\documentclass[review,12pt]{elsarticle}

\usepackage{amssymb}
\usepackage{amsmath}
\usepackage[utf8]{inputenc} 
\usepackage[T1]{fontenc}    
\usepackage{hyperref}      
\usepackage{url}           
\usepackage{booktabs}       
\usepackage{amsfonts}       
\usepackage{nicefrac}       
\usepackage{microtype}      
\usepackage{lipsum}		
\usepackage{graphicx}
\usepackage{natbib}
\usepackage{doi}
\usepackage{amsmath}
\usepackage{amssymb}
\usepackage{booktabs}
\usepackage{listings}
\usepackage{color}
\usepackage{algorithm} 
\usepackage{multirow}
\usepackage{colortbl}
\usepackage{listings}
\usepackage{makecell}
\usepackage{xcolor}
\usepackage{tabularray}
\usepackage[normalem]{ulem}
\usepackage{cuted}
\usepackage[inkscapelatex=false]{svg}
\usepackage{subfig}

\journal{Pattern Recognition}

\begin{document}

\begin{frontmatter}



\title{Adaptively Bypassing Vision Transformer Blocks for Efficient Visual Tracking}


\author[1]{Xiangyang Yang}
\author[2]{Dan Zeng}
\author[1]{Xucheng Wang}
\author[1]{You Wu}
\author[1]{Hengzhou Ye}
\author[3]{Qijun Zhao}
\author[1]{Shuiwang Li\corref{cor1}}
\ead{lishuiwang0721@163.com}
\cortext[cor1]{Corresponding author}

\affiliation[1]{organization={College of Computer Science and Engineering},
            addressline={Guilin University of Technology}, 
            city={Guilin},
            postcode={541004}, 
            country={China}}
\affiliation[2]{organization={Department of Computer Science and Engineering},
            addressline={Southern University of Science and Technology}, 
            city={Shenzhen},
            postcode={518055}, 
            country={China}}
\affiliation[3]{organization={College of Computer Science},
            addressline={Sichuan University}, 
            city={Chengdu},
            postcode={610065}, 
            country={China}}

%% Abstract
\begin{abstract}
Empowered by transformer-based models, visual tracking has advanced significantly. However, the slow speed of current trackers limits their applicability on devices with constrained computational resources. To address this challenge, we introduce ABTrack, an adaptive computation framework that adaptively bypassing transformer blocks for efficient visual tracking. 
The rationale behind ABTrack is rooted in the observation that semantic features or relations do not uniformly impact the tracking task across all abstraction levels. Instead, this impact varies based on the characteristics of the target and the scene it occupies.
Consequently, disregarding insignificant semantic features or relations at certain abstraction levels may not significantly affect the tracking accuracy. We propose a Bypass Decision Module (BDM) to determine if a transformer block should be bypassed, which adaptively simplifies the architecture of ViTs and thus speeds up the inference process. To counteract the time cost incurred by the BDMs and further enhance the efficiency of ViTs, we introduce a novel ViT pruning method to reduce the dimension of the latent representation of tokens in each transformer block. Extensive experiments on multiple tracking benchmarks validate the effectiveness and generality of the proposed method and show that it achieves state-of-the-art performance. Code is released at: \href{https://github.com/xyyang317/ABTrack}{https://github.com/xyyang317/ABTrack}.
\end{abstract}


% \begin{highlights}
% \end{highlights}

%% Keywords

\end{frontmatter}

\section{Introduction}
\label{sec:intro}
Visual object tracking is a foundational task in computer vision, involving the continuous monitoring of an object's movement within a video sequence from its initial state \cite{Li2018DeepVT}. Over the past decade, significant advancements in deep neural networks have propelled this field forward. Networks such as AlexNet \cite{AlexNet}, ResNet~\cite{ResNet}, GoogLeNet~\cite{googlenet}, alongside the recent integration of transformers \cite{2017Attention}, have collectively enhanced the accuracy and robustness of tracking systems. Transformers, in particular, have been pivotal in the development of high-performance trackers \cite{TransT,Stark,wang2021transformer,mixformer,ostrack}, offering substantial improvements in tracking precision and reliability.
Despite these technological leaps, there is a noticeable gap in the balance between performance and efficiency. Many state-of-the-art tracking methods are optimized for top-tier performance on powerful GPUs, often at the expense of tracking speed \cite{TransT,SiameseRPN,ToMP,CSWinTT}. These methods, while achieving real-time performance on high-end hardware, frequently struggle to maintain the same level of efficiency on resource-constrained devices. This highlights the need for further research to develop solutions that are both accurate and efficient, making them adaptable to various hardware capabilities.

\begin{figure*}[t]
	\centering
\includegraphics[width=1\textwidth]{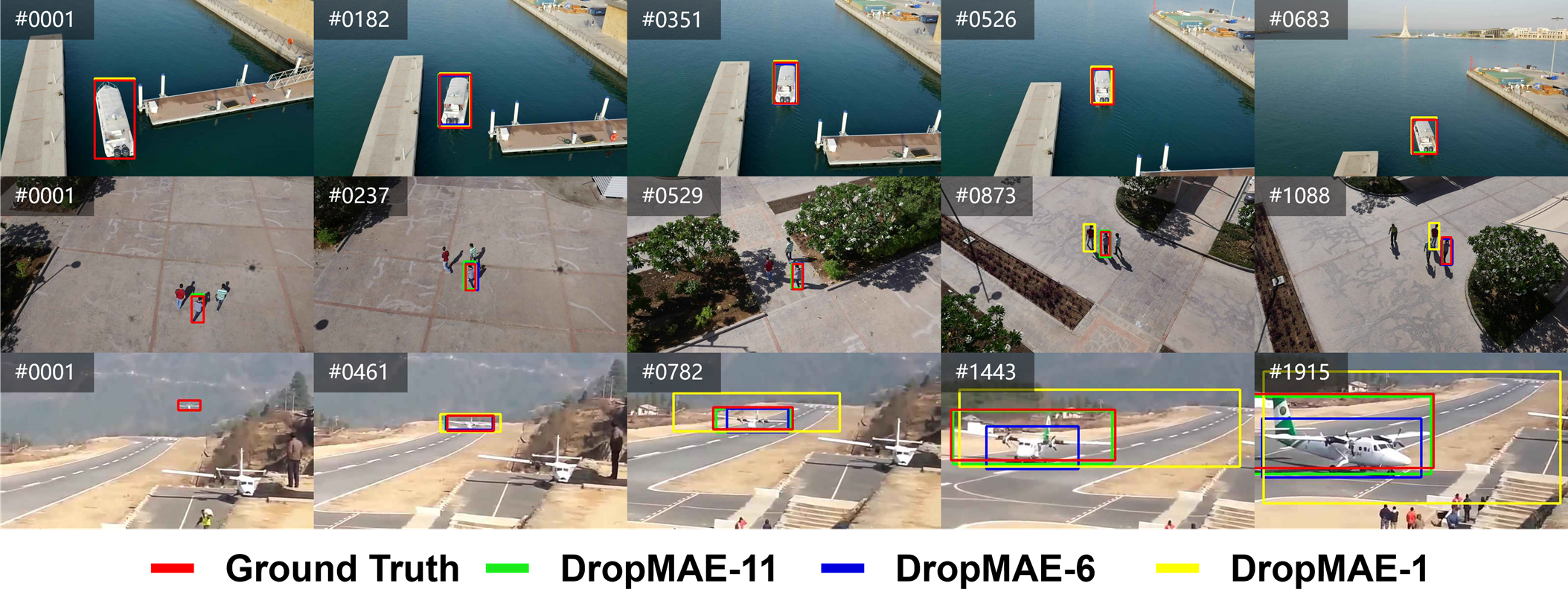}
	\caption{This figure shows the tracking results of three trackers adapted from DropMAE~\cite{Wu2023DropMAEMA} with different number of ViT layers, dubbed DropMAE-$N$, $N= 1, 6, 11$ being the number of layers. Note that despite DropMAE-11 successfully tracking all three targets, DropMAE-1 succeeds in tracking the boat with only one layer and DropMAE-6 succeeds in tracking the boat and the person with only six layers. These results suggest that \textbf{deep semantic features or relations are not always necessary for visual tracking task.} } \label{fig:motivation}
\end{figure*}

Recently, the tracking community has observed a surge in the adoption of single-stream architectures, where feature extraction and fusion are seamlessly integrated via harnessing the capabilities of pre-trained ViT backbone networks~\cite{ostrack,simtrack,mixformer}. In our study, we also employ a single-stream architecture, but our focus lies in enhancing the efficiency of ViTs for more effective visual tracking. The attention mechanism in ViTs transforms image patch representation vectors repeatedly, progressively incorporating semantic relationships between image patches through a series of transformer blocks~\cite{ViT,liu2021swin}. This capability empowers ViTs to extensively capture both local features or relations within objects and broader-scale ones between objects and objects/background, thereby significantly enhancing their performance in image classification tasks~\cite{wang2021pyramid,Arnab2021ViViTAV,mehta2021mobilevit}. However, a critical distinction exists between visual tracking and image classification, which necessitates further investigation. 
In image classification, the local and broader-scale features or relationships essential for capturing class characteristics are intrinsic to the object and remain unaffected by surrounding circumstances. In contrast, visual tracking requires a dynamic understanding of semantic features and relationships. The significance of these features is intricately tied to the specific characteristics of both the target and the surrounding environment.
In other words, the tracking process is not solely dictated by a fixed set of semantic features or relations across all levels of abstraction. Instead, it operates as a dynamic and context-sensitive mechanism, involving continuous adaptation and interpretation of dynamic changes in the object's surroundings. This dynamic nature of tracking necessitates a flexible and context-sensitive approach, where the tracker must dynamically adjust its focus and attention based on the evolving context of the scene. For instance, in scenarios where a target moves against a monochromatic background, simply the color contrast between the target and the background can be sufficient for effective tracking. However, when the target is moving in a cluttered environment, capturing adequate semantic features and relations becomes crucial for achieving effective tracking (see Figure \ref{fig:motivation} for an intuitive illustration). This observation underscores the need for a tracking methodology that adapts to the intricacies of each tracking scenario, recognizing that the significance of semantic features or relations is context-dependent. 
Unfortunately, this insightful perspective has been largely neglected in visual tracking research. More comprehensive explorations could be both fruitful and worthwhile, potentially leading to significant advancements in the field.

In this work, we propose an adaptive computation framework for efficient visual tracking based on the above observation. Given that the difficulty of tracking a target depends on the background it occupies, we propose to adaptively bypass certain transformer blocks. This allows for the selective extraction of significant semantic features or relations tailored to specific tracking tasks, while disregarding blocks representing insignificant, negligible or repeated features or relations. To achieve this, we introduce a bypass decision module into each transformer block. This module outputs a bypass probability, which determines whether a transformer block should be bypassed. 
In our implementation, the bypass decision module is a linear layer followed by a nonlinear activation function, which takes an introduced bypass token as input. 
By tailoring the architecture of ViTs to the specific demands of a tracking task, our approach holds the potential to accelerate the inference process for visual tracking. To counteract the computational burden incurred by these introduced modules and further improve the efficiency of ViTs, we also reduce the dimension of latent representation of tokens by introducing a novel ViT pruning method. 
This innovative adjustment includes a local ranking of importance scores and an extended applicability due to our generalized formulation. 
We called this proposed adaptive computation framework ABTrack. Extensive experiments substantiate the effectiveness and generality of our method and demonstrate that our ABTrack achieves state-of-the-art real-time performance.
Our contributions can be summarized as follows:
\begin{itemize}
    \item In recognition of that semantic features or relations may not uniformly impact the tracking task across all levels of abstraction, we introduce the Bypass Decision Module and the block sparsity loss. This module enables the adaptive bypassing of transformer blocks to enhance efficiency for visual tracking.
    \item To counteract the time cost of BDM and further enhance efficiency, we propose to reduce the dimension of latent representations of tokens by introducing a novel ViT pruning method based on local ranking.
    \item We introduce ABTrack, a family of efficient trackers based on these components,  which demonstrates promising performance while maintaining extremely fast tracking speeds. The proposed components can be integrated seamlessly with other ViT-based trackers. Extensive experiments validate the effectiveness of the proposed method and show that ABTrack achieves state-of-the-art real-time performance.
\end{itemize}

The rest of this paper is organized as follows. Section \ref{section_related_work} reviews related research. Section \ref{section_method} details the methodology of the proposed technique. Section \ref{section_experiment} describes the experiments conducted and discusses the results. Lastly, Section \ref{section_conclusion} presents the conclusions and insights drawn from the study.

\section{Related Work}\label{section_related_work}
\label{sec:relatedwork}
\subsection{Visual Tracking}
Siamese-based approaches, such as SiamFC~\cite{SiameseFC} and SiamRPN~\cite{SiameseRPN} are widely used in visual tracking.
This framework typically utilizes two backbone networks sharing parameters to extract features from the template and search region images. It employs a correlation-based network to interact with the features and uses head networks for prediction. To further enhance tracking performance, TransT~\cite{TransT}, TrSiam~\cite{wang2021transformer}, and their subsequent works such as Stark~\cite{Stark}, ToMP~\cite{ToMP}, CSWinTT~\cite{CSWinTT}, and GRM \cite{Gao2023GeneralizedRM} introduce the transformer for feature interaction. Recently, a single-stream framework, exemplified by such as MixFormer~\cite{mixformer}, OSTrack~\cite{ostrack}, SimTrack~\cite{simtrack}, DropTrack~\cite{Wu2023DropMAEMA}, CTTrack~\cite{Song_2023_AAAI}, and ARTrack~\cite{wei2023autoregressive}, has demonstrated great success in visual tracking. This framework combines feature extraction and fusion within a single backbone network. It is a simple yet effective approach that capitalizes on the capabilities of pre-trained image classification models. However, these methods were primarily designed for powerful GPUs, resulting in slow speeds on edge devices, which limits their practical applications.
In this work, we also adopt this single-stream framework but with a focus on improving its efficiency.
This emphasis on efficiency aims to make the framework more suitable for deployment on resource-constrained devices, enhancing its practical utility across various real-world applications.

\subsection{Efficient Tracking Methods}
Practical applications necessitate the use of efficient tracking systems capable of delivering exceptional performance at rapid speeds, especially when deployed on edge devices. While early techniques like ECO~\cite{danelljan2017eco} and ATOM~\cite{ATOM} were able to achieve real-time tracking on edge devices, their performance fell short in comparison to state-of-the-art trackers available today. Recent advancements have led to the development of more efficient tracking solutions. For instance, 
LightTrack~\cite{yan2021lighttrack} employs neural architecture search (NAS) to discover network configurations, resulting in a low computational overhead and relatively high performance. FEAR~\cite{borsuk2022fear}  introduces a family of efficient and accurate tracking methods through the utilization of a dual-template representation and a pixel-wise fusion block. Recently, 
HiT~\cite{Kang2023ExploringLH} presents a family of efficient transformer-based tracking models that use a bridge module and dual-image position encoding to bridge the gap between lightweight transformers and the tracking framework. 
MixFormerV2 ~\cite{mixformerv2} proposes a fully transformer-based tracking framework without any dense convolutional operations or complex score prediction modules, additionally,  it employs a distillation-based model compression paradigm to enhance its efficiency.
DyTrack~\cite{Dytrack} introduces a dynamic transformer framework that automatically configures reasoning paths based on the complexity of the input to make better use of computational resources.
Despite significant advancements, these methods are highly tailored and have problems with universality. A widely applicable method that can be easily integrated into existing trackers is more desirable. In this work, we present a novel approach that is straightforward and can be seamlessly integrated into existing ViT-based methods.

\subsection{Efficient Vision Transformers}

ViT~\cite{ViT} revolutionized the field of computer vision with its impressive performance~\cite{touvron2021training,wang2021pyramid,liu2021swin}. However, despite their powerful modeling capabilities, they have been facing challenges associated with speed limitations.
Especially on resource-constrained edge devices, the demand for efficient ViTs intensifies. 
To expedite ViTs, various lightweight ViTs have been recently introduced, employing techniques like low-rank methods, model compression, and hybrid designs~\cite{Zhang2022MiniViTCV,Mao2021TPruneET,Li2022EfficientFormerVT}. However, low-rank and quantization-based ViTs often sacrifice considerable accuracy for efficiency, while Hybrid ViTs with CNN-based components exhibit restricted input flexibility. In this work, we introduce a novel ViT pruning method based on local ranking to reduce the dimension of the latent representation of tokens to boost efficiency. 
On the other hand, recent endeavors in efficient ViTs based on conditional computation have explored adaptive inference methods for accelerated modeling. For instance, DynamicViT~\cite{Rao2021DynamicViTEV} introduces extra control gates to pause specific tokens using the Gumbel-softmax trick. 
DVT~\cite{DVT} recognizes that different images require varying numbers of tokens, and it automatically configures an appropriate number of tokens for each input image to improve computational efficiency.
A-ViT~\cite{yin2022vit} uses an Adaptive Computation Time (ACT)-like method to enhance efficiency, accuracy, and token importance allocation without requiring additional halting sub-networks. 
TPS~\cite{Wei2023JointTP} proposes a joint Token Pruning and Squeezing module for compressing vision transformers with higher efficiency.
However, variable number of tokens in these methods incurs considerable time cost due to additional unstructured assess operations. Very recently, LGViT~\cite{LGViT} accelerates the inference process of ViT through a dynamic early exiting mechanism, which is a more structured approach employing dynamic ViT blocks rather than varying the token number. Nevertheless, its training process is highly complex, and it cannot trim non-adjacent ViT blocks due to the constraints of the early exiting mechanism.
In this study, we investigate adaptively bypassing ViT blocks as a structured and more effective method to enhance ViTs' efficiency. As the tracking challenge for a specific target is highly context-dependent, adaptive bypassing of ViT blocks can notably decrease insignificant semantic features or relations at specific abstraction levels without significantly impacting the tracker's performance.

\section{Proposed Approach}\label{section_method}
\label{sec:method}
\begin{figure*}[t]
	\centering
\includegraphics[width=1\textwidth]{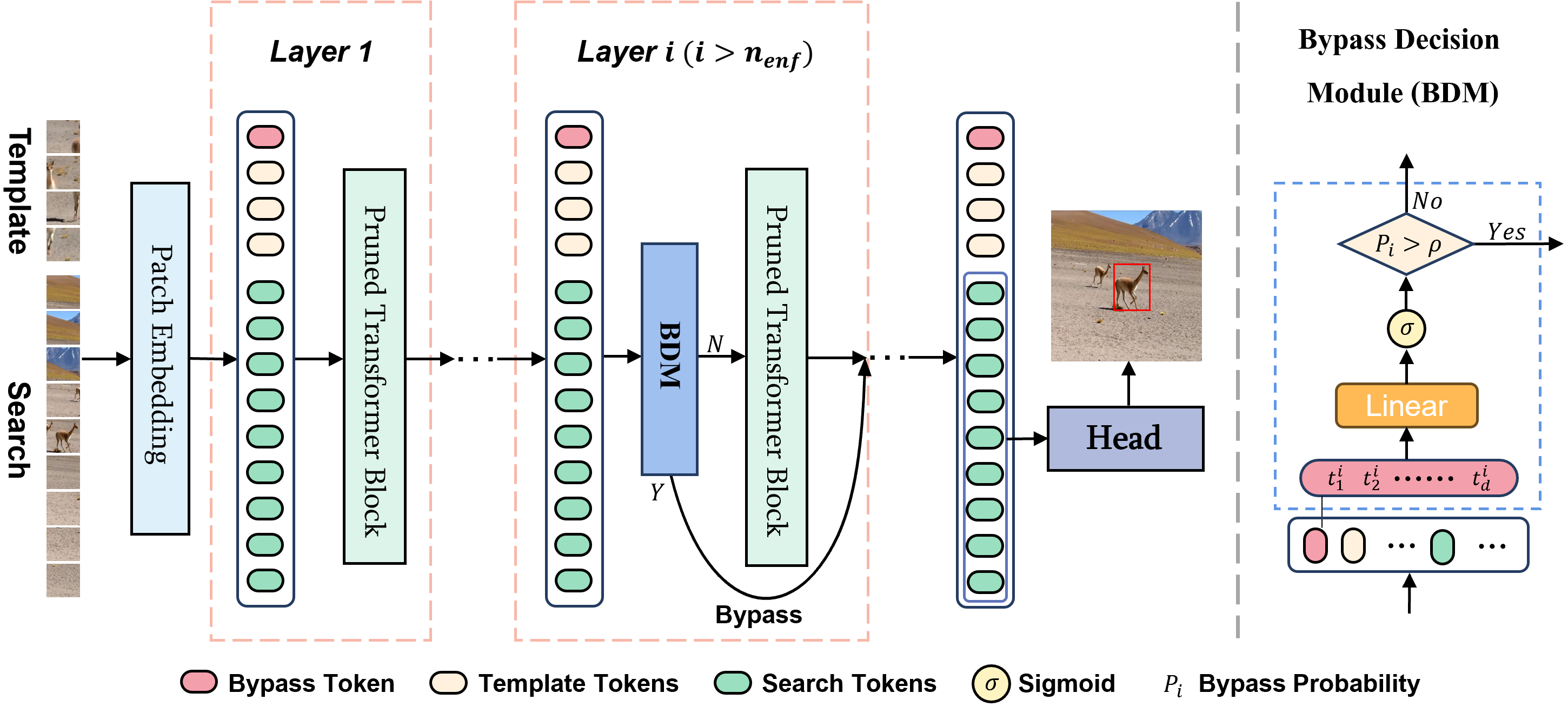}
	\caption{ \textbf{(Left)} Overview of the proposed ABTrack framework. It contains a single-stream backbone and a prediction head, in which the backbone consists of pruned ViT blocks and Bypass Decision Modules. \textbf{(Right)} The implementation details of the proposed Bypass Decision Module. } \label{fig:overview}
\end{figure*}

In this section, we begin with a brief overview of our tracking framework, ABTrack, as illustrated in Figure \ref{fig:overview}. Next, we introduce the bypass decision module and the approach to pruning vision transformer models. Finally, we provide details on the prediction head and training objective.

\subsection{Overview}

The proposed ABTrack is a single-stream tracking framework, consisting of an adaptive ViT-based backbone and a prediction head. The inputs to the framework are the target template and the search image. The extracted features from the backbone are fed into the prediction head to obtain tracking predictions. In the backbone network, we introduce a bypass decision module (BDM) for each ViT block (layer) except for the first $n_{enf}$ layers. Each BDM is trained to adaptively determines whether to bypass the associated block based on the computed bypassing probability. 
Furthermore, we conduct pruning on the ViT blocks using two introduced dimension reduction (DR) matrices. These matrices indicate the importance scores for each dimension of the linear layer of the Multi-Head Self-Attention (MSA) and Multi-Layer Perceptron (MLP) within each block, respectively.
In our approach, the BDM maintains the innate transformation characteristics of each ViT layer to ensure consistent information processing throughout the network. However, ViT pruning modifies the BDM's input domain, directly influencing its operational efficiency and performance. Therefore, we prioritize ViT pruning before integrating the BDMs, aiming to improve overall tracking accuracy and efficiency in visual tasks.
Finally, with BDMs integrated, the pruned model undergoes fine-tuning within the tracking framework.

\subsection{Bypass Decision Module (BDM)}
The objective of the BDMs is to dynamically determine whether a transformer block should be bypassed based on the given template and search image. In consideration of efficiency, it is implemented by a linear layer followed by a nonlinear activation function. Its input is a bypass token, a specific summary token that is appended to the sequence of image tokens. And its output represents the probability of bypassing the associated transformer block. Specifically, take the $i_{th}$ ($i>n_{enf}$) layer for example, let the bypass token be denoted by $(t^i_1, ..., t^i_d):=\textbf{b}^i\in \mathbb{R}^{d}$, 
the linear layer by $\mathfrak{l}^i$, where $d$ is the token dimension, then formally the BDM at layer $i$ is defined by
\begin{equation}
    p_i=\sigma(\mathfrak{l}^i(\textbf{b}^i)),
\end{equation}
where $p_i\in [0,1]$ represents the probability of bypassing the $i$ transformer block, $\sigma(x)=1/(1+e^{-x})$ denotes the sigmoid function.
If $p_i>\rho $, where $\rho \in [0, 1]$ is the bypass probability threshold, the transformer block at layer $i$ is bypassed; otherwise, the output tokens from the $(i-1)_{th}$ layer is fed into the $i_{th}$ transformer block. Let $N$ denote the total number of transformer blocks of the given ViT. 
Theoretically, all $N$ blocks could be bypassed simultaneously, resulting in no correlation being computed between the template and the search image. To prevent such situations, the first  $n_{enf}$ layers are enforced without bypassing. These low-level blocks are crucial because they provide foundational information that supports the construction of more complex features and representations. By prioritizing low-level features, the system ensures that the basic, yet critical, elements of the data are processed sufficiently, thus facilitating the subsequent development of higher-level abstractions. This practice also alleviates the computational burden caused by an excessive number of BDMs. 
Another extreme case occurs when few or no blocks are bypassed regardless of the input. This situation enhances the model's representation ability, allowing it to capture more detailed and complex features. Consequently, this increases the model's capacity to reduce training loss effectively. However, this can lead to higher computational costs and may not be efficient for real-time applications. 
Balancing between bypassing and retaining blocks is crucial to optimize both performance and efficiency. 
To this end, we propose an adaptive block sparsity loss, denoted as $\mathcal{L}_{spar}$.
This loss function adaptively penalizes smaller average bypass probabilities across all layers considered in the model. The penalty is stricter when the tracking task is deemed easy and less strict when the task is challenging. This adaptive approach dynamically adjusts block bypassing, aiming to strike a balance between efficiency and accuracy tailored to the complexity of each tracking scenario.
The block sparsity loss is formally defined by 
\begin{equation}
\begin{split}
    &\mathcal{L}_{spar}=|
    \frac{1}{N-n_{enf}}\sum^{N}_{i=n_{enf}+1}p_i-\tau(\mathcal{L}_{iou};\overline{\mathcal{L}_{iou}})|,\\
    &\tau(\mathcal{L}_{iou};\overline{\mathcal{L}_{iou}})=\text{clip}(\tau_0+\zeta(\mathcal{L}_{iou}-\overline{\mathcal{L}_{iou}}),0,1),
\end{split}
\end{equation}
where $\text{clip}(x, a, b)$ is a mathematical operation that limits the value of 
$x$ within a specified range defined by $a$ and 
$b$. $\tau \in [0,1]$ is a function of GIoU loss \cite{Rezatofighi2019GeneralizedIO} (see Section \ref{Head_Training}) and is used along with $\rho$ to control the block sparsity. Note that a linear function of the deviation of GIoU loss from its average is used to quantify the difficulty of the tracking task. In our implementation, the average GIoU loss, i.e., $\overline{\mathcal{L}_{iou}}$, is estimated with a batch training samples. $\zeta>0$ and $\tau_0\in [0,1]$ are two constants used to model the linear relation between $\tau$ and $(\mathcal{L}_{iou}-\overline{\mathcal{L}_{iou}})$. We call $\tau$ the block sparsity function. Given $\rho$, generally the larger $\tau$ is, the more sparse the model will be.

\subsection{Vision Transformer Pruning (VTP)}\label{section_VTP}
\begin{figure}
    \centering
    \subfloat[global ranking pruning]{
        \includegraphics[scale=0.55]{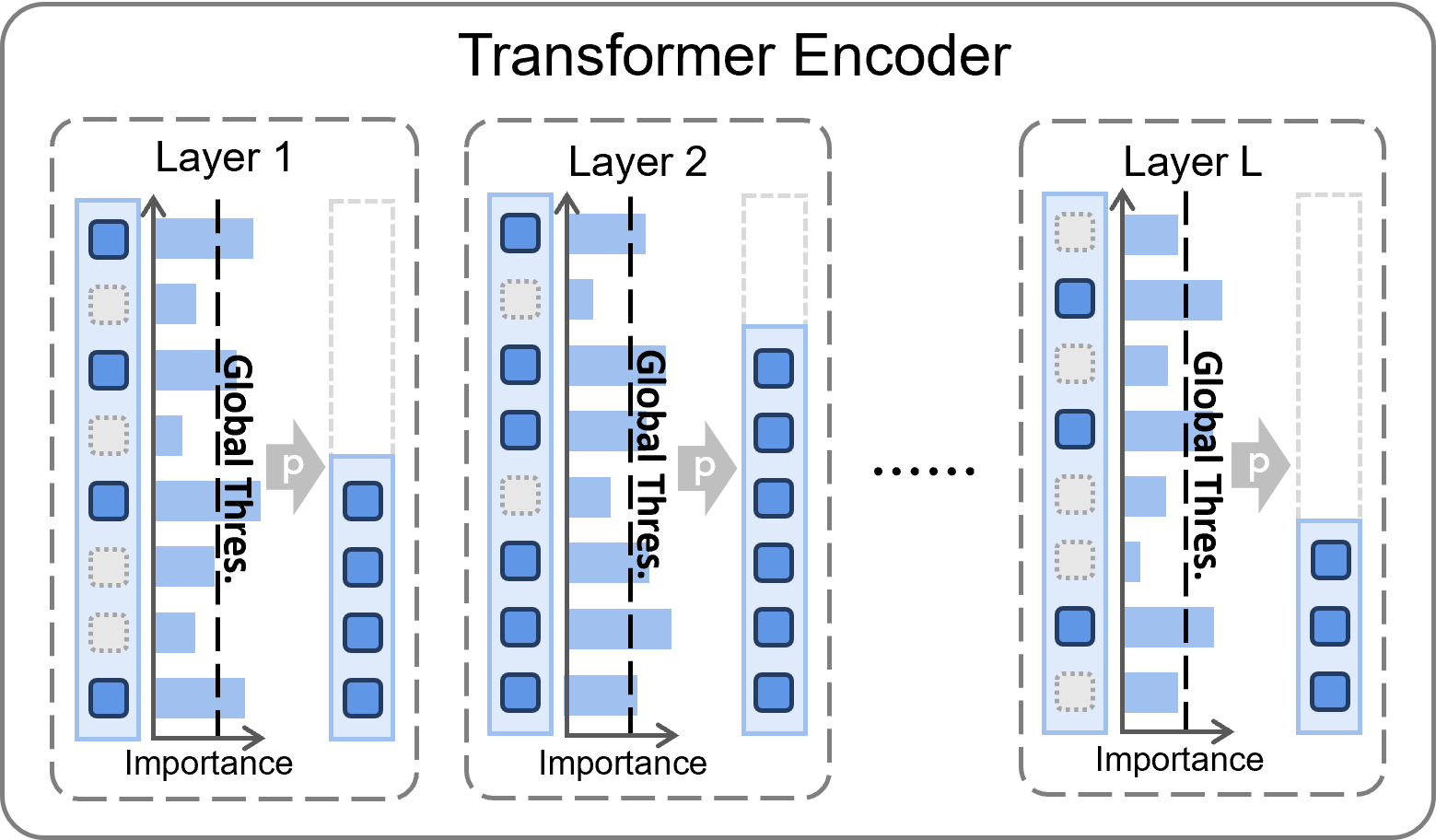}
    }
    \subfloat[local ranking pruning]{
        \includegraphics[scale=0.55]{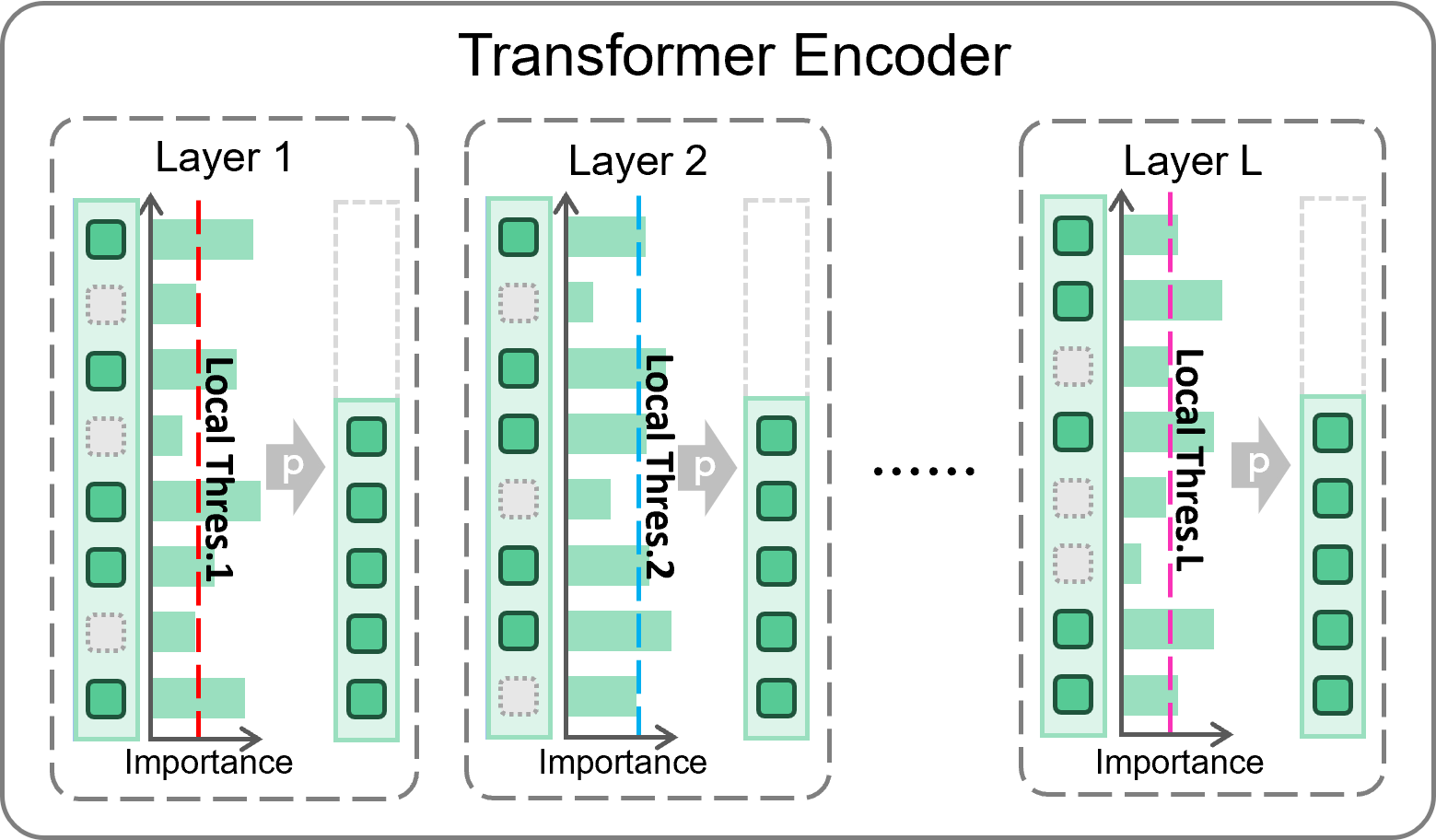}
    }
    \caption{Comparison between global ranking pruning and local ranking pruning, where the grey blocks represent the dimensions pruned in each transformer block.}
    \label{fig:pruning}
\end{figure}

While the architecture of our proposed Bypass Decision Module (BDM) is straightforward, it introduces a computational burden that cannot be ignored. To tackle this issue, we refine the ViT pruning method introduced in \cite{zhu2021vision} to minimize additional time costs and enhance efficiency. Our enhancements broaden the scope of the original method, initially designed for scaled dot-product attention mechanisms, to encompass ViTs employing multi-headed self-attention (MSA). 
Additionally, we employ a local ranking of importance scores within each ViT block, departing from the original global approach, as demonstrated in Figure \ref{fig:pruning}.  
In the original approach, as shown in Figure \ref{fig:pruning} (a), pruning thresholds were determined based on the average scores across all ViT blocks. However, this method could lead to under-pruning in layers with higher overall importance scores and over-pruning in layers with lower scores. Our proposed ViT pruning method, as shown in Figure \ref{fig:pruning} (b), evaluates the impact of dimensions within each layer locally, selectively pruning dimensions with low importance scores in each block.
These improvements not only extend the applicability of our proposed method to widely used ViTs but also address the limitations associated with the global ranking of importance scores, which are detailed as follows. 

Let $\mathcal{X}\in \mathbb{R}^{d\times n}$ denote an input sequence of $n$ tokens of dimensions $d$, the proposed pruning-enabled ViT layer that map $\mathcal{X}$ to $\mathcal{Z}$ is defined as follows,
\begin{equation}
\small
\begin{split}
&\mathcal{Y}=\mathbf{\check{MSA}}(\mathbf{LN}(\mathcal{X}))+\mathcal{X},\\
&\mathcal{Z}=\mathbf{\check{MLP}}(\mathbf{LN}(\mathcal{Y}))+\mathcal{Y},
\end{split}
\end{equation}
where $\mathbf{LN}$ denotes layer normalization, $\mathbf{\check{MSA}}$ and $\mathbf{\check{MLP}}$ are generalized MSA and multi-layer perceptron (MLP). Suppose $\mathbf{\check{MSA}}$ is parameterized by $\textbf{W}_i^Q,\textbf{W}_i^K,\textbf{W}_i^V  \in \mathbb{R}^{d\times (d/N_h)}$, $\textbf{W}^O \in \mathbb{R}^{d\times d}$, for $i=1,...,N_h$, and $N_h\in \mathbb{N}$ (i.e., the number of heads that divides $d$). Additionally, let $\mathbf{\check{MLP}}$ be parameterized by $\textbf{W}_k^L\in \mathbb{R}^{d\times d}$ for $ k=1,...,N_l$, where $N_l\in \mathbb{N}$ represents the number of linear layers. $\mathbf{\check{MSA}}$ and $\mathbf{\check{MLP}}$ are formally defined as follows,
\begin{equation}
\small
\begin{split}
&\mathbf{\check{MSA}}(\mathcal{X};\{\textbf{W}^Q,\textbf{W}_i^K,\textbf{W}_i^V\}^{N_h}_{i=1},\textbf{W}^O)\\&=\text{Concat}(\{\mathbf{Attention}(\widetilde{\mathcal{X}}\widetilde{\textbf{W}}^Q_i,\widetilde{\mathcal{X}}\widetilde{\textbf{W}}^K_i,\widetilde{\mathcal{X}}\widetilde{\textbf{W}}^V_i)\}^{N_h}_{i=1})\widetilde{\textbf{W}}^O,\\
&\mathbf{\check{MLP}}(\mathcal{Y};\textbf{W}^L_{1},...,\textbf{W}^L_{{N_l}})=\phi_{{N_l}}(...(\phi_1(\widetilde{Y}\widetilde{\textbf{W}}^L_{1})...)\widetilde{\textbf{W}}^L_{{N_l}}),
\end{split}
\end{equation}
where $\text{Concat}$ performs the concatenation operation, $\widetilde{\mathcal{X}}=\mathcal{X}\textbf{D}_1$, $\widetilde{\textbf{W}}^O=\textbf{D}_1{\textbf{W}}^O$, $\widetilde{\mathcal{Y}}=\mathcal{Y}\textbf{D}_2$, and
\begin{equation}
\small
\begin{split}
&\textbf{Attention}(\textbf{Q},\textbf{K},\textbf{V}) = \textbf{Softmax}(\textbf{K}^T\textbf{Q}/\sqrt{d})\textbf{V},\\
&\widetilde{\textbf{W}}_i^J=\textbf{D}_1{\textbf{W}}_i^J\quad \forall  \quad J\in \{Q,K,V\}\quad \&\quad i\in \{1,...,N_h\}\\
&\widetilde{\textbf{W}}_k^L=\textbf{D}_2{\textbf{W}}_k^L\quad \forall \quad k\in \{1,...,N_l\},\quad 
    \end{split}
\end{equation}
in which $\textbf{Softmax}(\cdot)$ denotes the row-wise softmax normalization function, $\phi_i$ denotes the activation function at layer $i$ of $\mathbf{\check{MLP}}$, and $\textbf{D}_1, \textbf{D}_2\in \mathfrak{D}:=\{diag(\mathbf{v})|\mathbf{v}\in \{0,1\}^d, N_h\mid ||\mathbf{v}||_{1}\}$ are two diagonal matrices used to select or mask out dimensions, called dimension reduction (DR) matrices. $||\cdot||_1$ indicates the $l^1$ norm. If $\textbf{D}_1$ and $\textbf{D}_2$ are identity matrix, then $\mathbf{\check{MSA}}$ and $\mathbf{\check{MLP}}$ reduce to the traditional formulations of Multi-Head Self-Attention (MSA) and Multi-Layer Perceptron (MLP) as presented in prior works ~\cite{ViT,2017Attention}. 

The objective of our proposed ViT pruning method is to optimize $\textbf{D}_1$ and $\textbf{D}_2$ as trainable parameters for each ViT layer. By treating these matrices as trainable, we can dynamically adjust the pruning process during training, allowing for a more fine-tuned and effective reduction of computational complexity. This approach leverages the inherent flexibility of deep learning optimization techniques, enabling the model to learn which dimensions are most critical for preserving performance while minimizing unnecessary computations.
Considering optimization difficulties due to their discrete values, $\textbf{D}_1$ and $\textbf{D}_2$ are relaxed to real values for end-to-end optimization. Let $\hat{\textbf{D}}_1^k,\hat{\textbf{D}}_2^k\in \hat{\mathfrak{D}}:=\{diag(\mathbf{v})|\mathbf{v}\in \mathbb{R}^d \}$ denote the relaxed DR matrices at layer $k$ of the ViT to be pruned. To enforce sparsity of all DR matrices and thereby achieve a higher model pruning ratio, a $L_1$ regularization term is added to the training loss, which is 
\begin{equation}
\mathcal{L}_{reg}=\sum_{k=1}^N \alpha_k(||\hat{\textbf{D}}_1^k||_1+||\hat{\textbf{D}}_2^k||_1),
\end{equation}
where $\alpha_k$ is a sparsity hyper-parameter to control the sparsity at layer $k$. 
After the ViT pruning training, the relaxed DR matrices $\{\hat{\textbf{D}}_1^k,\hat{\textbf{D}}_2^k\}_{k=1}^N$  are binarized according to a predefined pruning ratio $\mu$. This process involves setting the first $d^* = \lfloor d \times \mu / N_h \rfloor \times N_h$ values, ranked in descending order, to ones, and the remaining values to zeros. This binarization step ensures that the pruning is consistent with the desired level of sparsity, enabling efficient model performance while retaining the most critical features.
The pipeline for pruning is summarized as follows: 1) training with the sparsity regularization; 2) pruning dimensions of the parameter matrices with DR matrices; 3) fine-tuning.

\subsection{Prediction Head and Training Objective}\label{Head_Training}
As in \cite{mixformer, ostrack}, 
we employ a prediction head that consists of multiple Conv-BN-ReLU layers, to directly estimate the bounding box of the target. The output tokens associated with the search image are initially transformed into a 2D spatial feature map before being fed into the prediction head. The prediction head produces 
a target classification score $\mathbf{p} \in [0,1]^{H_x/P\times W_x/P}$, a local offset $\mathbf{o}\in [0,1]^{2\times H_x/P\times W_x/P}$, and a normalized bounding box size 
	$\mathbf{s} \in [0,1]^{2\times H_x/P\times W_x/P}$, where $H_x$ and $W_x$ respectively represent the height and width of the search image, while $P$ represents the patch size.  The initial estimate of the target position is determined by identifying the location with the highest classification score, i.e., $(x_c, y_c)=\textup{argmax}_{(x,y)}\mathbf{p}(x,y)$, based upon which the final target bounding box is estimated by 
	\begin{equation}
		\{(x_t,y_t);(w,h)\}=\{(x_c, y_c)+\mathbf{o}(x_c,y_c);\mathbf{s}(x_c,y_c)\}.
	\end{equation}
For the tracking task, we adopt the weighted focal loss $\mathcal{L}_{cls}$ \cite{Law2018CornerNetDO} for classification, a combination of the $L_1$ loss and the GIoU loss  $\mathcal{L}_{iou}$ \cite{Rezatofighi2019GeneralizedIO} for bounding box regression. Finally, the overall loss function is:
        \begin{equation}
		\begin{split}
			\mathcal{L}_{overall}=\mathcal{L}_{cls}+\lambda_{iou}\mathcal{L}_{iou}+\lambda_{L_1}\mathcal{L}_{L_1}+\gamma \mathcal{L}_{spar} ,
		\end{split}
	\end{equation}
	where the constants $\lambda_{iou}=2$ and $\lambda_{L_1}=5$ are the same as in \cite{mixformer,ostrack}, $\gamma$ is set to 5. Our framework is trained end-to-end with the overall loss $\mathcal{L}_{overall}$, after loading the pre-trained weights of the ViT for image classification.

\section{Experiments}
\label{section_experiment}

In this section, we extensively evaluate our method on five tracking benchmarks, including comparison with state-of-the-art methods, comprehensive ablation studies, and an analysis of generality.

\subsection{Implementation Details}
We employ A-ViT-T~\cite{yin2022vit}, DeiT-Tiny\cite{touvron2021training}, and ViT-Tiny\cite{touvron2021training} as the backbones to build our trackers: ABTrack-AViT, ABTrack-DeiT, and ABTrack-ViT. 
We use the training splits from several datasets for our training process, including GOT-10k \cite{2021GOT}, LaSOT \cite{Fan2018LaSOTAH}, COCO \cite{2014Microsoft}, and TrackingNet  \cite{2018TrackingNet}. 
In our experiments, we set the bypass probability threshold $\rho$, block sparsity parameter $\tau_0$, $\zeta$, pruning ratio $\mu$, and regularization sparsity hyper-parameter $\alpha_k$ to 0.5, 0.4, 0.1, 0.3, and $ 10 ^ {-4}$, respectively. These parameters were carefully chosen to balance the trade-offs between model complexity, computational efficiency, and tracking performance.
We use the AdamW optimizer \cite{Loshchilov2017DecoupledWD} with a weight decay of $ 10 ^ {-4} $ for finally training the model, the initial learning rate is set to $ 4 \times 10 ^ {-5} $, and the models are trained for 300 epochs with a batch size of 32. The learning rate is reduced by a factor of 10 after 240 epochs. 

\subsection{Datasets}
\textbf{GOT-10k} \cite{2021GOT}
 is a large-scale benchmark for generic object tracking in the wild.  It contains over 10,000 video sequences with diverse challenges.
\textbf{TrackingNet} \cite{2018TrackingNet}
 is a high-quality benchmark for object tracking.  It consists of 30,000 video sequences with precise pixel-level annotations, covering a wide range of object categories and tracking scenarios.
\textbf{LaSOT} \cite{Fan2018LaSOTAH}
 is a comprehensive benchmark for single object tracking.  It contains 1,400 video sequences with a total of 3.5 million frames, covering various challenges.
\textbf{UAV123} \cite{Mueller2016ABA}
is a large-scale benchmark for aerial tracking, comprising 123 sequences with over 112K frames. It covers a variety of challenging aerial scenarios. 
 \textbf{VOT-ST2021} \cite{vot2021} is a visual tracking challenge that tests algorithms on short-term, single object tracking in RGB images. 

\setlength{\parskip}{0.2cm plus4mm minus3mm}
\subsection{State-of-the-art Comparisons}
Our trackers are extensively compared with 13 state-of-the-art trackers on four tracking benchmarks, including 6 real-time and 7 non-real-time trackers. 
Note that the inference speeds are evaluated on GOT-10k, and we classify trackers with speeds exceeding 20 FPS on all three platforms as real-time, while others are considered non-real-time. The three platforms used are Nvidia TitanX GPU, Intel i7-12700k (3.6GHz) CPU, and Nvidia Jetson AGX Xavier edge device. The experimental results are presented in Table \ref{tab:sota}. 
\begin{table*}[t]
\centering
\caption{Compared with the state-of-the-art methods on the GOT-10k \cite{2021GOT}, LaSOT \cite{Fan2018LaSOTAH}, TrackingNet \cite{2018TrackingNet} and UAV123~\cite{Mueller2016ABA} \protect\footnotemark. The top three real-time results are displayed in {\textbf{\color[HTML]{FE0000}Red}}, {\textbf{\color[HTML]{3531FF}blue}} and {\textbf{\textcolor[rgb]{0,0.502,0}{green}}} fonts, while the best non-real-time results are in \textbf{bold} font.}
\label{tab:sota}
\resizebox{5.4in}{!}{
\begin{tabular}{c|c|c|ccc|ccc|ccc|cc|ccc} 
\hline
\multirow{2}{*}{} & \multirow{2}{*}{Method} & \multirow{2}{*}{Source} & \multicolumn{3}{c|}{GOT-10k}                                                                                                      & \multicolumn{3}{c|}{LaSOT}                                                                                                        & \multicolumn{3}{c|}{TrackingNet}                                                                                                  & \multicolumn{2}{c|}{UAV123}                                                           & \multicolumn{3}{c}{Speed(FPS)}                                                                                                \\
                  &                         &                         & AO                                        & SR$_{0.5}$                                   & SR$_{0.75}$                                  & AUC                                       & P$_{Norm}$                                   & P                                         & AUC                                       & P$_{Norm}$                                   & P                                         & AUC                                       & P                                         & GPU                                      & CPU                                     & AGX                                      \\ 
\hline
\multirow{9}{*}{\rotatebox{90}{Real-time}} & \textbf{ABTrack-AViT}            & \multirow{3}{*}{\textbf{Ours}}   & \textcolor{red}{\textbf{66.8}}            & \textcolor{red}{\textbf{77.1}}            & \textcolor{red}{\textbf{59.6}}            & \textcolor{red}{\textbf{63.4}}            & \textcolor{red}{\textbf{73.3}}            & \textcolor{red}{\textbf{67.1}}            & \textcolor{red}{\textbf{79.3}}            & \textcolor{red}{\textbf{84.4}}            & \textcolor{red}{\textbf{76.1}}            & \textcolor{red}{\textbf{65.2}}            & \textcolor{red}{\textbf{82.3}}            & \textcolor[rgb]{0,0.502,0}{\textbf{195}} & \textcolor[rgb]{0,0.502,0}{\textbf{52}} & 53                                       \\
                  & \textbf{ABTrack-DeiT}            &                         & \textcolor[rgb]{0,0.502,0}{\textbf{64.0}} & \textcolor[rgb]{0,0.502,0}{\textbf{74.2}} & \textcolor[rgb]{0,0.502,0}{\textbf{56.1}} & \textcolor{blue}{\textbf{61.1}}           & \textcolor{blue}{\textbf{70.1}}           & \textcolor{blue}{\textbf{62.8}}           & \textcolor{blue}{\textbf{78.1}}           & \textcolor{blue}{\textbf{83.2}}           & \textcolor{blue}{\textbf{74.3}}           & \textcolor{blue}{\textbf{64.4}}           & \textcolor[rgb]{0,0.502,0}{\textbf{81.8}} & \textcolor{blue}{\textbf{223}}           & \textcolor{blue}{\textbf{64}}           & \textcolor{blue}{\textbf{65}}            \\
                  & \textbf{ABTrack-ViT}             &                         & 61.9                                      & 72.2                                      & 53.0                                      & 59.9                                      & 68.8                                      & 60.8                                      & 77.2                                      & \textcolor[rgb]{0,0.502,0}{\textbf{82.5}} & 73.1                                      & \textcolor[rgb]{0,0.502,0}{\textbf{64.3}} & \textcolor{blue}{\textbf{81.9}}           & \textcolor{red}{\textbf{241}}            & \textcolor{red}{\textbf{69}}            & \textcolor{red}{\textbf{68}}             \\ 
\cline{2-17}
                  & MixFormerV2~\cite{mixformerv2}             & NIPS23                  & 61.9                                      & 71.7                                      & 51.3                                      & \textcolor[rgb]{0,0.502,0}{\textbf{60.6}} & \textcolor[rgb]{0,0.502,0}{\textbf{69.9}} & 60.4                                      & 75.8                                      & 81.1                                      & 70.4                                      & 63.7                                      & 81.3                                      & 167                                      & 45                                      & 49                                       \\
                  & HiT~\cite{Kang2023ExploringLH}                      & ICCV23                  & 62.6                                      & 71.2                                      & 54.4                                      & 60.5                                      & 68.3                                      & 61.5                                      & \textcolor[rgb]{0,0.502,0}{\textbf{77.7}} & 81.9                                      & 73.1                                      & 63.3                                      & 80.5                                      & 173                                      & 47                                      & \textcolor[rgb]{0,0.502,0}{\textbf{59}}  \\
                  & E.T.Track~\cite{Blatter2021EfficientVT}               & WACV23                  & 56.6 		                                         & 64.6                                        & 42.5                                         & 59.1                                      & 66.8	 	                                         & 60.1                                        & 72.5                                      & 77.8                                      & 69.5                                      & 62.3                                      & 81.5                                      & 56                                       & 32                                     & 22                                       \\
                  & FEAR~\cite{borsuk2022fear}                     & ECCV22                  & 61.9                                      & 72.2                                      & 52.5                                         & 53.5                                      & 59.7                                        & 54.5                                      & 70.2                                      & 80.8                                      & 71.5                                      & 58.9                                      & 79.2                                      & 119                                      & 34                                      & 38                                       \\
                  & TCTrack~\cite{Cao2022TCTrackTC}                 & CVPR22                  & \textcolor{blue}{\textbf{66.2}}           & \textcolor{blue}{\textbf{75.6}}           & \textcolor{blue}{\textbf{61.0}}             & 60.5                                      & 69.3                                      & \textcolor[rgb]{0,0.502,0}{\textbf{62.4}} & 74.8                                      & 79.6                                      & \textcolor[rgb]{0,0.502,0}{\textbf{73.3}} & 60.5                                      & 80.0                                        & 140                                      & 45                                      & 41                                       \\
                  & LightTrack~\cite{yan2021lighttrack}               & CVPR21                  & 61.1                                      & 71.0                                        & 54.3                                         & 53.8                                      & 60.5                                         & 53.7                                      & 72.5                                      & 77.8                                      & 69.5                                      & 59.9                                      & 77.6                                      & 134                                      & 39                                      & 36                                       \\ 
\hline
\multirow{7}{*}{\rotatebox{90}{Non-real-time}} & ARTrack~ \cite{wei2023autoregressive}                 & CVPR23                  & \textbf{73.5}                             & 82.2                                      & \textbf{70.9}                             & \textbf{70.4}                             & \textbf{79.5}                             & \textbf{76.6}                             & \textbf{84.2}                             & \textbf{88.7}                             & \textbf{83.5}                             & 67.9                                      & 85.9                                      & 26                                       & 9                                       & 8                                        \\
                  & GRM~\cite{Gao2023GeneralizedRM}                     & CVPR23                  & 73.4                                      &82.9                           & 70.4                                      & 69.9                                      & 79.3                                      & 75.8                                      & 84.0                                        & \textbf{88.7}                             & 83.3                                      & 67.7                                      & 85.4                                      & 53                                       & \textbf{18}                             & 16                                       \\
                  & CTTrack~\cite{Song_2023_AAAI}                 & AAAI23                  & 71.3                                      & 80.7                                      & 70.3                                      & 67.8                                      & 77.8                                      & 74.0                                      & 82.5                                      & 87.1                                      & 80.3                                      & 68.8                                      & 89.5                                      & 40                                       & 13                                       & 12                                        \\
                  & OSTrack~\cite{ostrack}                 & ECCV22                  & 70.5                                      & 80.9                                      & 66.7                                      & 69.1                                      & 78.7                                      & 75.2                                      & 81.9                                      & 87.2                                      & 81.0                                        & 67.5                                      & 84.7                                      & \textbf{68}                              & \textbf{18}                             & \textbf{19}                              \\
                  & CSWinTT~\cite{CSWinTT}                 & CVPR22                  & 69.4                                      & 78.9                                      & 65.4                                      & 66.2                                      & 75.2                                      & 70.9                                      & 81.9                                      & 86.7                                      & 79.5                                      & \textbf{70.5}                             & \textbf{90.3}                             & 10                                       & 3                                       & 3                                        \\
                  & ToMP~\cite{ToMP}                    & CVPR22       		             & 72.2                                        & \textbf{83.9}                                         & 65.5                                       & 68.5                                      & 79.2                                      & 73.5                                      & 81.5                                      & 86.4                                      & 78.9                                      & 68.5                                      & 87.3                                      & 24                                       & 8                                       & 7                                        \\
                  & TrSiam~\cite{wang2021transformer}                  & CVPR21                  & 67.3                                      & 78.7                                      & 58.6                                      & 62.4                                      & 70.5                                         & 60.6                                      & 78.1                                      & 82.9                                      & 72.7                                      & 67.4                                      & 85.3                                         & 55                                       & 7                                       & 10                                       \\
\hline
\end{tabular}}
\end{table*}
\footnotetext{Some methods did not report results on certain datasets in their original papers. We have supplemented these results in our table by testing the official code.}
In terms of tracking performance, our ABTrack-AViT surpasses all other real-time trackers across all metrics, while our ABTrack-DeiT secures the second position in most metrics on the four test benchmarks. For instance, on TrackingNet, ABTrack-AViT and ABTrack-DeiT take the first and second place respectively, demonstrating superior tracking performance compared to MixFormerV2 with more than 2\% gain across all three performance metrics.
Regarding speed, our trackers demonstrate the fastest speeds across all three platforms, except for ABTrack-AViT on AGX, which is slightly slower than HiT. 
Although our trackers may demonstrate tracking performance inferior to most non-real-time trackers, their exceptional speed sets them apart. Specifically, only  ABTrack-DeiT and ABTrack-ViT achieve GPU speeds exceeding 200 FPS, and all our trackers surpass the leading ARTrack \cite{wei2023autoregressive} by more than 7 times across all speed metrics. Additionally, compared to TrSiam~\cite{wang2021transformer}, a tracker with the closest tracking performance, our ABTrack-AViT achieves over 3 times the GPU speed, over 7 times the CPU speed, and over 5 times the AGX speed. These results underscore the effectiveness of our methodology and its state-of-the-art performance.

\begin{table}
\centering
\caption{Comparison with real-time trackers on VOT-ST2021~\cite{vot2021}.}
\label{tab:vot}
\resizebox{5.4in}{!}{
\begin{tabular}{c|ccccc|ccc} 
\toprule
Method     & E.T.Track~\cite{Blatter2021EfficientVT} & LightTrack~\cite{yan2021lighttrack} & FEAR~\cite{borsuk2022fear}                                      & HiT~\cite{Kang2023ExploringLH}  & MixFormerV2~\cite{mixformerv2}                              & \textbf{ABTrack-ViT}                                & \textbf{ABTrack-DeiT}                     & \textbf{ABTrack-AViT}                      \\ 
\hline
EAO        & 0.224     & 0.225      & \textcolor[rgb]{0,0.502,0}{\textbf{0.250}} & 0.232 & 0.234                                      & 0.235                                      & \textcolor{red}{\textbf{0.279}}  & \textcolor{blue}{\textbf{0.277}}  \\
Accuracy   & 0.372     & 0.391      & 0.436                                      & 0.439 & \textcolor[rgb]{0,0.502,0}{\textbf{0.443}} & 0.428                                      & \textcolor{blue}{\textbf{0.445}} & \textcolor{red}{\textbf{0.449}}   \\
Robustness & 0.631     & 0.641      & 0.655                                      & 0.643 & 0.637                                      & \textcolor[rgb]{0,0.502,0}{\textbf{0.658}} & \textcolor{red}{\textbf{0.752}}  & \textcolor{blue}{\textbf{0.742}}  \\
\bottomrule
\end{tabular}}
\end{table}

The VOT-ST challenge series is widely recognized in the tracking community,  is a standard for tracking algorithm assessment. We additionally conducted a comparison with five real-time trackers on the VOT-ST2021 benchmark~\cite{vot2021}, and the results are presented in Table \ref{tab:vot}. 
As can be seen, our ABTrack-AViT and ABTrack-DeiT demonstrate superior performance compared to other real-time trackers, achieving Expected Average Overlap (EAO), Accuracy, and Robustness scores of 0.277, 0.449, and 0.742 for ABTrack-AViT, and 0.279, 0.445, and 0.752 for ABTrack-DeiT, respectively.

\begin{table}
\centering
\caption{Ablation study on the impact of the proposed Bypass Decision Module (BDM) and ViT Pruning (VTP).}
\label{tab:baseline}
\resizebox{4.2in}{!}{
\begin{tabular}{cccccccc} 
\toprule
        & BDM & VTP & P    & AUC  & FPS & FLOPs(G) & Params.(M)  \\ 
\hline
\multirow{3}{*}{ABTrack-AViT} &     &     & 83.8 & 66.4 & 181 & 2.39    & 8.27     \\
  &   $\checkmark$   &     & \textbf{84.1}~$_{0.3}$$\uparrow$ & \textbf{67.4}~$_{1.0}$$\uparrow$ & 225~$_{24.3\%}$$\uparrow$ & 0.97-2.39    & 3.52-8.27       \\
  &   $\checkmark$   &   $\checkmark$   &   82.3~$_{1.5}$$\downarrow$   &   65.2~$_{1.2}$$\downarrow$   &   \textbf{248}~$_{37.0\%}$$\uparrow$  & 0.88-1.88   & 3.31-6.36       \\
\bottomrule
\end{tabular}}
\end{table}

\subsection{Ablation Study}
\textbf{Impact of the Bypass Decision Module (BDM) and the ViT Pruning (VTP).} We add the proposed components BDM and VTP into the baseline (i.e., ABTrack-AViT without incorporating the proposed components) one by one to evaluate their effectiveness. The evaluation results on UAV123 are shown in Table \ref{tab:baseline}. 
As can be seen, the inclusion of BDM leads to a significant increase in speed.  Specifically, ABTrack-AViT experiences an increase of 24.3\% in GPU speed. For FLOPs and Params., it is evident that the model's computational load and parameters decrease after the implementation of the BDM. Notably, the Precision (P) and AUC also increase by 0.3\% and 1.0\%, respectively, which may be attributed to the enhanced generalization ability of the simplified model. These substantial improvements of efficiency and the improvement of accuracy underscore the effectiveness of our BDM. When the VTP is further integrated, ABTrack-AViT sees an additional 12.7\% increase in GPU speed, with only marginal drops of 1.2\% and 0.2\% in P and AUC respectively. This improvement in GPU speeds, coupled with minimal compromises in tracking performance, confirms the efficacy of our VTP in optimizing tracking efficiency.

\begin{table}[]
\begin{minipage}[t]{0.38\textwidth}
\centering
\makeatletter\def\@captype{table}
\caption{Comparison of time overheads of proposed components.}
\label{tab:time}
\resizebox{1.55in}{!}{
\begin{tabular}{cc} 
\toprule
Module        & Time(ms)  \\ 
\hline
ViT Block(VB) & 0.41      \\
VB+BDM        & 0.46      \\
VB+BDM+VTP    & 0.37      \\
\bottomrule
\end{tabular}}
\end{minipage}
\begin{minipage}[t]{0.05\textwidth}
\ 
\end{minipage}
\begin{minipage}[t]{0.56\textwidth}
\centering
\makeatletter\def\@captype{table}
\caption{Ablation study on the effect of the block sparsity function.}
\label{tab:implementations}
\resizebox{3.0in}{!}{
\begin{tabular}{c|cc|ccc} 
\toprule
\multirow{2}{*}{}                                 & \multicolumn{2}{c|}{UAV123}   & \multicolumn{3}{c}{GOT-10k}                    \\
                                                  & P             & AUC           & AO          & SR$_{0.5}$      & SR$_{0.75}$      \\ 
\hline
\multicolumn{1}{l|}{ABTrack-DeiT($\tau=\tau_0$)} & 78.2          & 61.3          & 60.8          & 71.2          & 52.3           \\
\multicolumn{1}{l|}{ABTrack-DeiT}                 & \textbf{81.8} & \textbf{64.4} & \textbf{64.0} & \textbf{74.2} & \textbf{56.1}  \\
\bottomrule
\end{tabular}}
\end{minipage}
\end{table}

\textbf{Time overheads of proposed components.}
We delve into the computational time overheads incurred by the proposed components to study their impact on the efficiency of our method. We calculate the time cost of a ViT Block (VB) and the time costs after the BDM and the VTP being integrated successfully. Note that these time costs, as shown in Table \ref{tab:time}, are computed by averaging 10,000 inferences to obtain stable measures. 
As can be seen, the time overhead of the BDM is about $12\%\approx (0.46-0.41)/0.41$ of the ViT block, and with the VTP integrated the time cost of the BDM is well offset and remarkably reduce about $9.7\%\approx (0.41-0.37)/0.41$ time cost of the VB.
These results suggest that bypassing a VB leads to approximately an 88\% reduction in the time cost associated with that block. Conversely, if a VB is not bypassed, the time cost is still reduced by about 9.7\% due to the VTP. These findings underscore the always enhancement in efficiency achieved by the proposed method.

\textbf{Effect of the block sparsity function.}
To study the effect of the proposed block sparsity function $\tau$ on learning adaptive bypassing ViT blocks, we train ABTrack-DeiT with $\tau$ being substituted by its initial value $\tau_0$ for comparison, which is denoted by ABTrack-DeiT($\tau=\tau_0$). The tracking performance of the two methods on UAV123 and GOT-10k are shown in Table \ref{tab:implementations}. 
It's evident that ABTrack-DeiT significantly outperforms ABTrack-DeiT($\tau=\tau_0$) on both datasets. Specifically, ABTrack-DeiT achieves gains of 3.6\% and 3.1\% in Precision (P) and AUC on UAV123, and 3.2\%, 3.0\%, and 3.8\% in AO, SR$_{0.5}$, and SR$_{0.75}$ on GOT-10k respectively. These results strongly support the effectiveness of the proposed block sparsity function for learning adaptively bypassing ViT blocks.

\begin{table}[]
\centering
\caption{Ablation study of the block sparsity constant $\tau_0$ on the P, AUC, and GPU speed of ABTrack on UAV123. }
\label{tab:jump}
\resizebox{2.8in}{!}{
\begin{tabular}{cccccccccc} 
\toprule
$\tau_0$ & 0.1                            & 0.2  & 0.3                                       & 0.4                             & 0.5  & 0.6  & 0.7                                      & 0.8                            & 0.9                            \\ 
\hline
P        & \textbf{\textcolor{red}{81.4}} & 80.7 & \textbf{\textcolor[rgb]{0,0.502,0}{80.8}} & \textbf{\textcolor{blue}{80.9}} & 79.3 & 79.1 & 78.8                                     & 78.5~                          & 77.8~                          \\
AUC      & \textbf{\textcolor{red}{63.8}} & 62.5 & \textcolor[rgb]{0,0.502,0}{\textbf{62.6}} & \textcolor{blue}{\textbf{63.1}} & 62.1 & 62.4 & 61.9                                     & 61.5~                          & 61.1                           \\
FPS      & 223                            & 235  & 243                                       & 266                             & 286  & 307  & \textbf{\textcolor[rgb]{0,0.502,0}{315}} & \textbf{\textcolor{blue}{319}} & \textbf{\textcolor{red}{325}}  \\
\hline
\end{tabular}}
\end{table}

\textbf{Impact of the block sparsity constant.}
To investigate the impact of the block spasity constant $\tau_0$ on performance, we trained ABTrack-DeiT with values of $\tau_0$ ranging from 0.1 to 0.9 and evaluated it on the UAV123. The results are shown in Table \ref{tab:jump}. Overall, with an increase in the block sparsity constant $\tau_0$,  the accuracy decrease is small, while the speed improvement is substantial. For example, when $\tau_0$ increases from 0.1 to 0.9, the P and AUC respectively decrease by 3.6\% and 2.7\%, but the speed increases by a remarkable 45.7\%. This justifies that our BDM is capable of significantly improving efficiency without causing substantial deterioration in tracking performance, which is useful for deploying ViT-based trackers to platforms with limited computing resources.
Noticeable, within the 0.2-0.4 range, there is an observed improvement in accuracy, suggesting that choosing an appropriate $\tau_0$ can lead to a more favorable trade-off between speed and accuracy.

\begin{table}[]
\centering
\caption{Ablation study on when to initiate the BDM. Only layers following the layer $n_{enf}$ are incorporated with BDM.}
\label{tab:n_enf}
\resizebox{3.2in}{!}{
\begin{tabular}{ccccccccccc} 
\toprule
$n_{enf}$     & 0                             & 1                              & 2                                         & 3    & 4    & 5    & 6                                         & 7    & 8                              & 9                                \\ 
\hline
P           & 82.0~                         & 81.6                           & ~82.9~                                    & 82.3 & 82.5 & 82.8 & \textcolor[rgb]{0,0.502,0}{\textbf{83.0}} & 82.8 & \textcolor{red}{\textbf{83.4}} & \textcolor{blue}{\textbf{83.2}}  \\
AUC         & 64.2                          & ~64.0                          & \textcolor[rgb]{0,0.502,0}{\textbf{65.2}} & 64.7 & 64.9 & 65.0 & \textcolor[rgb]{0,0.502,0}{\textbf{65.2}} & 65.1 & \textcolor{red}{\textbf{65.7}} & \textcolor{blue}{\textbf{65.5}}  \\
FPS         & \textcolor{red}{\textbf{287}} & \textcolor{blue}{\textbf{276}} & \textbf{\textcolor[rgb]{0,0.502,0}{266}}  & 253  & 246  & 235  & 221                                       & 210  & 193                            & 181                              \\
avg. Layers & 7.2                           & 7.6                            & 8.0                                         & 8.4  & 8.8  & 9.2  & 9.6                                       & 10.0   & 10.4                           & 10.8                             \\
\bottomrule
\end{tabular}}
\end{table}

\textbf{Study on when to initiate the BDMs. }
To investigate the impact of the timing of initiating BDMs on performance, we trained ABTrack-DeiT with $n_{enf}$ ranging from 0 to 9 and evaluated on UAV123. The results are shown in Table~\ref{tab:n_enf}. It can be observed that as $n_{enf}$ increases, P and AUC generally exhibit an upward trend, while FPS decreases.
At $n_{enf}=2$, both AUC and FPS rank third, with P ranking fourth. However, when compared to the top performance of 83.4\% P and 65.7\% AUC at $n_{enf}=8$, its P and AUC are only 0.5\% lower, while the speed is 73 FPS higher. Although the speed decreases by 10 FPS when compared to $n_{enf}=1$, the increase in P and AUC is 1.3\% and 1.2\%, respectively.
This illustrates a nuanced balance between speed and accuracy, showing that $n_{enf}=2$ provides a good compromise, significantly boosting speed without a substantial sacrifice in performance.

\textbf{Impact of global and local ranking on ViT pruning.}
In this section, we analyze the impact of global and local ranking methods on the pruning effectiveness of ViTs. The objective is to understand how these two strategies affect the model's performance and efficiency. The impact of global and local ranking strategies on ViT pruning was assessed using the UAV123 and GOT-10k datasets. The evaluation results, presented in Table \ref{tab:pruning_method}, reveal that the local ranking strategy significantly outperforms the global ranking strategy.
Specifically, the local ranking strategy achieves a 1.6\% and 1.7\% improvement over the global ranking strategy in Precision (P) and AUC, respectively, on UAV123. On GOT-10k, it leads to improvements of 1.9\% in AO, 2.2\% in SR$_{0.5}$, and 1.8\% in SR$_{0.75}$.
These results indicate that the local ranking method is more effective for pruning ViT models, leading to better performance across multiple metrics on both datasets. This superiority is likely due to the ability of local ranking to more accurately identify and retain the most critical features within each layer, whereas global ranking may overlook layer-specific importance, resulting in less optimal pruning decisions.

\begin{table}[]
\centering
\caption{Impact of global and local ranking on ViT pruning.}
\label{tab:pruning_method}
\resizebox{3.2in}{!}{
\begin{tabular}{c|c|cc|ccc} 
\toprule
\multirow{2}{*}{}             & \multirow{2}{*}{ranking strategy} & \multicolumn{2}{c|}{UAV123}   & \multicolumn{3}{c}{GOT-10K}                    \\
                              &                                   & P             & AUC           & AO          & SR$_{0.5}$      & SR$_{0.75}$      \\ 
\hline
\multirow{2}{*}{ABTrack-DeiT} & global ranking                            & 80.2          & 62.7          & 62.1          & 72.0          & 54.3           \\
                              & local ranking                            & \textbf{81.8} & \textbf{64.4} & \textbf{64.0} & \textbf{74.2} & \textbf{56.1}  \\
\bottomrule
\end{tabular}}
\end{table}

\subsection{Study on the Generality}
To demonstrate the generality of our proposed approach, we integrated it into multiple ViT backbones and state-of-the-art trackers. We then evaluated the performance on the UAV123 dataset \cite{Mueller2016ABA}. This evaluation highlights the versatility and effectiveness of our method across different models. All hyperparameters were set to their default values. 

\textbf{Application to backbone of various scales.} We use three scales of DeiT as backbones, i.e., DeiT-Tiny, DeiT-Small, and DeiT-Base, to study the effectiveness of the proposed method on different scales of backbones. To make sense of each component's effect, they are integrated into the baselines separately. The experimental results are shown in Table \ref{tab:backbones}. As observed, for all these backbones, whether with BDM or both components integrated, there is a substantial improvement in the GPU speed of the baselines, with an increase of over 20.0\%. Meanwhile, the reductions in P and AUC are marginal, not exceeding 2.0\%. 
For example, when integrating both components into the baseline with DeiT-Small as the backbone, the speed raises by 53.4\%, accompanied by only marginal decreases of 1.5\% in P and 1.6\% in AUC. It's noteworthy that integrating BDM into the baseline with DeiT-Tiny as the backbone results in a 0.2\% increase in P and a 0.1\% increase in AUC. Additionally, there is a 25.5\% increase in speed.
These results highlight the adaptability and effectiveness of the proposed method across various scales of ViT backbones, demonstrating its potential to enhance both speed and performance in visual tracking tasks.

\begin{table}[]
\begin{minipage}[t]{0.48\textwidth}
\makeatletter\def\@captype{table}
\caption{Application of BDM and VTP to backbone of various scales, i.e., DeiT-Base, DeiT-Small, and DeiT-Tiny. }
\label{tab:backbones}
\resizebox{2.6in}{0.85in}{
\begin{tabular}{ccccccc} 
\hline
Backbone                                & BDM          & VTP          & P                    & AUC                  & FPS (GPU)              \\ 
\hline
\multirow{3}{*}{DeiT-Base}  &              &              & \textbf{86.6~}       & \textbf{67.8~}       & 59.7~            \\
                            &                       $\checkmark$ &              & 85.7~$_{0.9}$$\downarrow$          & 67.1~$_{0.7}$$\downarrow$          & 71.7~$_{20.1\%}$$\uparrow$            \\
                            &                       $\checkmark$ & $\checkmark$ & 84.6~$_{2.0}$$\downarrow$          & 66.3~$_{1.5}$$\downarrow$          & \textbf{86.0}~$_{44.1\%}$$\uparrow$   \\ 
\hline
\multirow{3}{*}{DeiT-Small} &              &              & \textbf{83.6~}       & \textbf{66.1~}       & 152.9~           \\
                            &                       $\checkmark$ &              & 82.8~$_{0.8}$$\downarrow$          & 65.5~$_{0.6}$$\downarrow$          & 195.8~$_{28.1\%}$$\uparrow$           \\
                            &                       $\checkmark$ & $\checkmark$ & 82.1~$_{1.5}$$\downarrow$          & 64.5~$_{1.6}$$\downarrow$          & \textbf{234.6}~$_{53.4\%}$$\uparrow$  \\ 
\hline
\multirow{3}{*}{DeiT-Tiny}  &              &              & 82.7~                & 65.1~       & 211.9~           \\
                            &                       $\checkmark$ &              & \textbf{82.9}~$_{0.2}$$\uparrow$ & \textbf{65.2}~$_{0.1}$$\uparrow$ & 265.9~$_{25.5\%}$$\uparrow$           \\
                            &                       $\checkmark$ & $\checkmark$ & 81.8~$_{0.9}$$\downarrow$          & 64.4~$_{0.7}$$\downarrow$          & \textbf{288.3}~$_{36.1\%}$$\uparrow$  \\
\hline
\end{tabular}}
\end{minipage}
\begin{minipage}[t]{0.02\textwidth}
\,
\end{minipage}
\begin{minipage}[t]{0.48\textwidth}
\makeatletter\def\@captype{table}
\caption{Application of BDM and VTP to SOTA trackers, i.e., DropMAE \cite{Wu2023DropMAEMA}, GRM \cite{Gao2023GeneralizedRM} and OSTrack \cite{ostrack}.}
\label{tab:methods}
\resizebox{2.6in}{0.85in}{
\begin{tabular}{cccccc} 
\hline
Method                    & BDM          & VTP          & P                    & AUC                  & FPS (GPU)             \\ 
\hline
\multirow{3}{*}{DropMAE} &              &              & \textbf{89.8~}       & \textbf{70.9~}       & 17.6~           \\
                         & $\checkmark$ &              & 88.9~$_{0.9}$$\downarrow$          & 70.0~$_{0.9}$$\downarrow$          & 20.9~$_{18.8\%}$$\uparrow$           \\
                         & $\checkmark$ & $\checkmark$ & 87.6~$_{2.2}$$\downarrow$          & 69.0~$_{1.9}$$\downarrow$          & \textbf{23.5~}$_{33.5\%}$$\uparrow$  \\ 
\hline
\multirow{3}{*}{GRM}     &              &              & 85.4~                & 67.7~                & 52.8~           \\
                         & $\checkmark$ &              & 86.2~$_{0.8}$$\uparrow$          & 68.3~$_{0.6}$$\uparrow$          & 61.5~$_{16.5\%}$$\uparrow$           \\
                         & $\checkmark$ & $\checkmark$ & \textbf{86.8~}$_{1.4}$$\uparrow$ & \textbf{68.6~}$_{0.9}$$\uparrow$ & \textbf{68.7~}$_{30.1\%}$$\uparrow$  \\ 
\hline
\multirow{3}{*}{OSTrack} &              &              & \textbf{84.7~}       & \textbf{67.5~}       & 68.3~           \\
                         & $\checkmark$ &              & 84.2~$_{0.5}$$\downarrow$          & 66.9~$_{0.6}$$\downarrow$          & 81.7~$_{19.6\%}$$\uparrow$           \\
                         & $\checkmark$ & $\checkmark$ & 83.6~$_{1.1}$$\downarrow$          & 66.5~$_{1.0}$$\downarrow$          & \textbf{92.1~}$_{34.8\%}$$\uparrow$  \\
\hline
\end{tabular}}
\end{minipage}
\end{table}

\textbf{Application to state-of-the-art trackers.}
To show that our approach is general and can be easily integrated into any ViT-based tracking frameworks, we apply our method to three state-of-of-art trackers, i.e., DropMAE \cite{Wu2023DropMAEMA},  GRM \cite{Gao2023GeneralizedRM}, and OSTrack \cite{ostrack}. The two components are subsequently incorporated into these trackers for comparison. The experimental results are shown in Table \ref{tab:methods}. 
As can be seen, all baseline models experience a substantial increase in GPU speed, whether integrating BDM alone or both components together, with improvements exceeding 15.0\%. Particularly, integrating BDM alone enhances the GPU speed of OSTrack by 19.6\%. Moreover, with both components integrated, its speed further rises to 34.8\%.
Regarding tracking performance, we observe that the P and AUC of DropMAE and OSTrack experience only a slight decrease, while GRM unexpectedly shows an improvement. Specifically, when BDM is integrated, the P and AUC of DropMAE and OSTrack decrease by less than 1.0\%, while those of GRM increase by more than 0.5\%. Surprisingly, with both components integrated, the P of GRM is raised by 1.4\%. These results validate the applicability of our method to enhance the efficiency of existing ViT-based trackers without significantly compromising their tracking
performance. Moreover, it may even enhance the baseline's performance by reducing redundant parameters, thereby improving its generalization ability.

\begin{figure*}[!h]
	\centering
\includegraphics[width=0.9\textwidth]{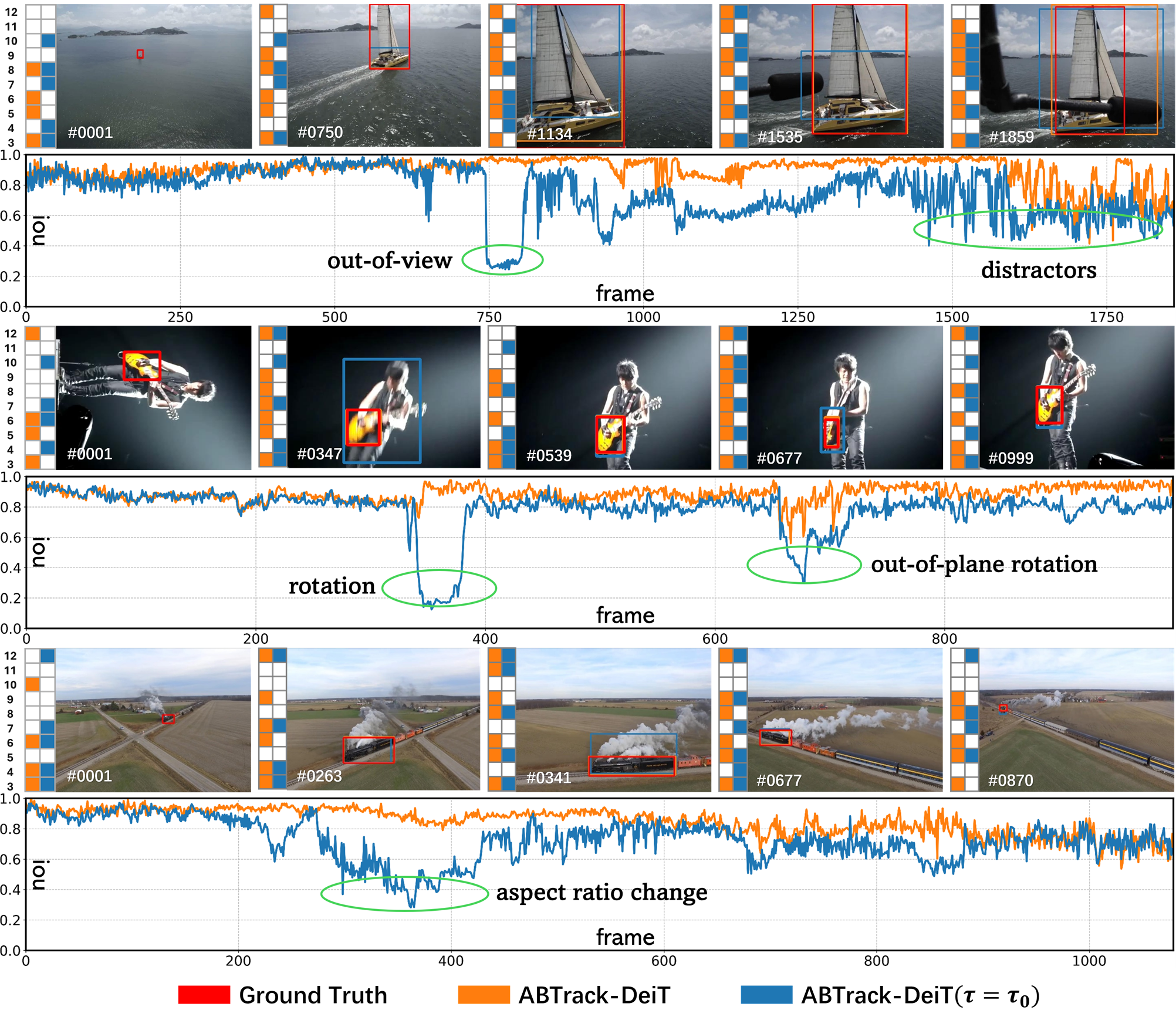}
	\caption{Illustration of the number of remaining ViT blocks, the IOU, and the predicted bounding boxes of ABTrack-DeiT and ABTrack-DeiT($\tau=\tau_0$) across three samples from LaSOT.} \label{fig:tracking_results}
\end{figure*}

\section{Qualitative Results}
\label{sec:visualization}

To make sense of the effectiveness of the proposed block sparsity loss in guiding the model to dynamically bypass ViT blocks according to the tracking task, we visualize in Figure \ref{fig:tracking_results} the number of remaining ViT blocks alongside the Intersection over Union (IOU) and the predicted bounding boxes of ABTrack-DeiT and ABTrack-DeiT($\tau=\tau_0$) across three samples from the LaSOT dataset \cite{Fan2018LaSOTAH}.
The color blocks in the vertical bars indicate retained ViT blocks, while those white blocks represent bypassed ViT blocks. The plots illustrate the variation of the IOU with respect to the frame for both methods. 
As oberved, when faced with challenging conditions such as out-of-view targets, distractors, changes in aspect ratio, and rotations, ABTrack-DeiT tends to retain more ViT blocks than ABTrack-DeiT($\tau=\tau_0$). This adaptive behavior suggests that the block sparsity loss effectively guides the model to preserve essential information for maintaining tracking accuracy under complex conditions.
Conversely, in scenarios where tracking conditions are less demanding and these challenges are absent, ABTrack-DeiT retains fewer ViT blocks than ABTrack-DeiT($\tau=\tau_0$). This efficiency-driven adaptation demonstrates the proposed block sparsity loss is able to reduce computational overhead by guiding the model to bypass unnecessary computations when simpler scenarios are encountered.
Moreover, ABTrack-DeiT demonstrates notably higher IOU scores under these challenging conditions, as indicated by the ellipses marked on the plots. These examples qualitatively emphasize the effectiveness and benefits of the proposed method, particularly the proposed block sparsity loss, in adaptively bypassing ViT blocks for visual tracking.

\begin{figure*}[t]
	\centering
\includegraphics[width=0.9\textwidth]{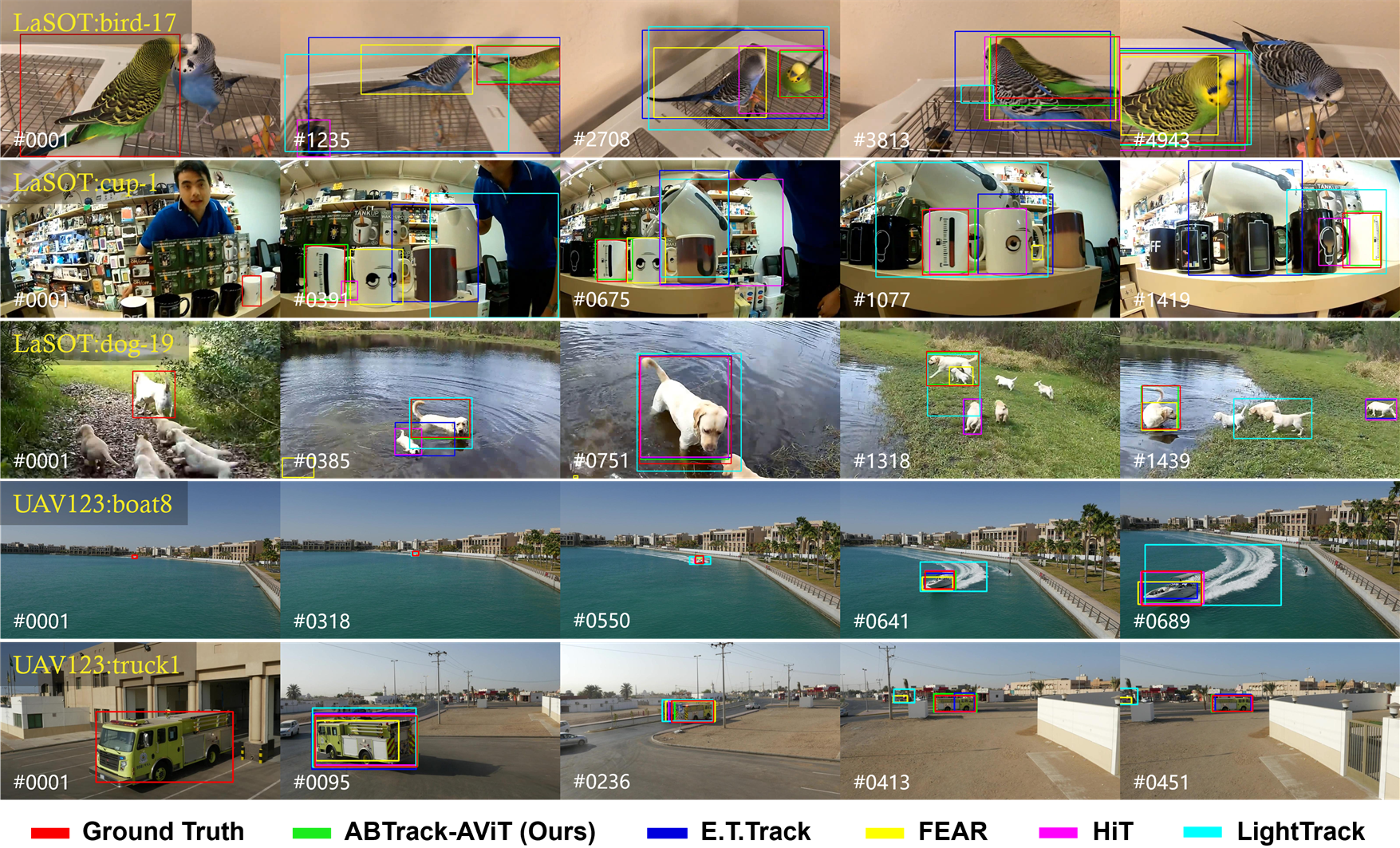}
	\caption{Qualitative analysis on five video sequences. These sequences were sourced from LaSOT and UAV123.} \label{fig:vis2}
\end{figure*}

Figure \ref{fig:vis2} presents a qualitative tracking analysis comparing our ABTrack-AViT with four state-of-the-art real-time trackers across five diverse video sequences. As illustrated, our tracker is the only one that consistently tracks the target in all challenging scenarios. These scenarios include scale variation (all examples), pose variation (bird-17, dog-19), similar objects (bird-17, cup1, dog-19), complex motion patterns (boat8), and occlusion (truck1). These results further underscore the effectiveness of our proposed approach.

\section{Conclusion}
\label{section_conclusion}
In this work, we delve into the efficiency of ViT-based visual tracking methods, highlighting a previously underappreciated  insight: the importance of semantic features or relationships at a specific abstraction level depends on the target and the contextual nuances of its environment.
We proposed a bypass decision module and a block sparsity loss to facilitate learning adaptively bypassing ViT blocks, thereby enhancing tracking efficiency. Additionally, we reduced the dimension of latent token representations by introducing a local ranking pruning strategy to further enhance efficiency. Our method can substantially enhance tracking speed without compromising considerable precision, unveil new opportunities for visual tracking applications with limited computational resources. Notably, our method is straightforward and seamlessly integrable into existing ViT-based tracking frameworks. Extensive experiments confirm its effectiveness, demonstrating that our ABTrack achieves state-of-the-art performance across multiple benchmarks.

\end{document}